\documentclass[twocolumn]{article}

\usepackage[utf8]{inputenc} 
\usepackage[T1]{fontenc}    
\usepackage{hyperref}       
\usepackage{url}            
\usepackage{booktabs}       
\usepackage{amsfonts}       
\usepackage{nicefrac}       
\usepackage{microtype}      
\usepackage{lipsum}
\usepackage{xcolor}
\usepackage{amsmath}

\usepackage{bm}
\usepackage{xparse}
\usepackage{tikz}
\usepackage{tkz-graph}
\usepackage{array,multirow,multicol,ctable}
\usetikzlibrary{arrows}
\usetikzlibrary{positioning}
\usepackage{enumerate}

\usepackage{comment}

\begin{document}
\title{Anomaly Detection at Scale: \\The Case for Deep Distributional Time Series Models}

\author{Fadhel Ayed, Lorenzo Stella, Tim Januschowski,  Jan Gasthaus \\
Amazon Research, \\
Berlin,\\
Germany\\
fadhel.ayed@gmail.com,  \{stellalo,  tjnsch, gasthaus \}@amazon.de}

\maketitle

\begin{abstract}
This paper introduces a new methodology for detecting anomalies in time series data, with a primary application to monitoring the health of (micro-) services and cloud resources. The main novelty in our approach is that instead of modeling time series consisting of real values or vectors of real values, we model time series of probability distributions over real values (or vectors). This extension to time series of probability distributions allows the technique to be applied to the common scenario where the data is generated by requests coming in to a service, which is then aggregated at a fixed temporal frequency. Our method is amenable to streaming anomaly detection and scales to monitoring for anomalies on millions of time series. We show the superior accuracy of our method on synthetic and public real-world data. On the Yahoo Webscope data set, we outperform the state of the art in 3 out of 4 data sets and we show that we outperform popular open-source anomaly detection tools by up to 17\% average improvement for a real-world data set.
\end{abstract}

\newcommand{\networkTrainingPicture}{%
 \begin{tikzpicture}[shorten >=1pt,->,draw=black!50, font=\footnotesize, scale=0.48]%
     \tikzstyle{every pin edge}=[<-,shorten <=1pt]
     \tikzstyle{node}=[circle,fill=gray!20,minimum size=25pt,inner sep=0pt]
     \tikzstyle{box}=[rectangle,draw=black!50,minimum height=12pt, inner sep=0pt]
     \tikzstyle{input node}=[draw=none, fill=none];
     \tikzstyle{network node}=[box, draw, fill=gray!20, minimum width=32pt];
     \tikzstyle{output node}=[box, minimum width=42pt,inner sep=2pt];
     \tikzstyle{projection node}=[box, minimum width=42pt,inner sep=2pt];
     \tikzstyle{spline node}=[box, minimum width=42pt,inner sep=2pt];
     \tikzstyle{loss node}=[box, minimum width=42pt,inner sep=2pt];
     \tikzstyle{node dots}=[node, fill=black, scale=0.2];
     \tikzstyle{annot} = [text width=4em, text centered]
 
      \node[draw=none, fill=none] (h0) at (1.5,1.8){$\mathbf{h}_{i, 0}$};
 
      \foreach \i/\j in {1/0, 2/1} {%
        \node[input node] (x\i) at (4*\i,0 ) {$z_{i,\j}, \mathbf{x}_{i, \i}$};%
        \node[network node] (h\i) at (4*\i,1.8) {$\mathbf{h}_{i, \i} | \Theta$};%
        \node[projection node] (p\i) at (4*\i,3.6) {$\theta_{i,\i} | \mathbf{h}_{i, \i}, \Phi$};%
        \node[spline node] (s\i) at (4*\i,5.2) {$q_{i,\i}(\alpha ; \theta_{i, \i})$};%
        \node[loss node] (l\i) at (4*\i,7) {$\mathcal{L}(q_{i,\i}, z_{i, \i})$};
        
        \path (x\i) edge (h\i);%
        \path (h\i) edge (p\i);%
        \path (p\i) edge (s\i);%
        \path (s\i) edge (l\i);%
      }%
 
      \node [left of=x1, node distance=1.05in, align=flush right] (xtlabel) {inputs};%
      \node [left of=h1, node distance=1.0in, align=flush right] (htlabel) {network};%
      \node [left of=p1, node distance=0.95in, align=flush right] (ptlabel) {projection};%
      \node [left of=s1, node distance=1.06in, align=flush right] (stlabel) {spline};%
      \node [left of=l1, node distance=1.0in, align=flush right] (ltlabel) {crps loss};%
 
      \node[input node] (d1) at (11.5, 1.8) {$\dots$};%
      \path (h0) edge (h1);%
      \path (h1) edge (h2);%
      \path (h2) edge (d1);%
 
      \foreach \c/\i/\j in {3/{t_0-1}/{t_0-2}, 4/{t_0}/{t_0-1}} {%
        \node[input node] (x\c) at (5*\c, 0) {$z_{i,\j}, \mathbf{x}_{i, \i}$};%
        \node[network node] (h\c) at (5*\c,1.8) {$\mathbf{h}_{i, \i} | \Theta$};%
        \node[projection node] (p\c) at (5*\c,3.6) {$\theta_{i,\i} | \mathbf{h}_{i, \i}, \Phi$};%
        \node[spline node] (s\c) at (5*\c,5.2) {$q_{i,\i}(\alpha ; \theta_{i, \i})$};%
        \node[loss node] (l\c) at (5*\c,7) {$\mathcal{L}(q_{i,\i}, z_{i, \i})$};
        
        \path (x\c) edge (h\c);%
        \path (h\c) edge (p\c);%
        \path (p\c) edge (s\c);%
        \path (s\c) edge (l\c);%
      }%
 
      \node[input node] (d2) at (23.5, 1.8) {$\dots$};%
      \path (d1) edge (h3);%
      \path (h3) edge (h4);%
      \path (h4) edge (d2);%
 	
      \foreach \c/\i/\j in {5/{T}/{T-1}} {%
        \node[input node] (x\c) at (5.3*\c,0 ) {$z_{i,\j}, \mathbf{x}_{i, \i}$};%
        \node[network node] (h\c) at (5.3*\c,1.8) {$\mathbf{h}_{i, \i} | \Theta$};%
        \node[projection node] (p\c) at (5.3*\c,3.6) {$\theta_{i,\i} | \mathbf{h}_{i, \i}, \Phi$};%
        \node[spline node] (s\c) at (5.3*\c,5.2) {$q_{i,\i}(\alpha ; \theta_{i, \i})$};%
        \node[loss node] (l\c) at (5.3*\c,7) {$\mathcal{L}(q_{i,\i}, z_{i, \i})$};
        
        \path (x\c) edge (h\c);%
        \path (h\c) edge (p\c);%
        \path (p\c) edge (s\c);%
        \path (s\c) edge (l\c);%
      }%
      
      \path (d2) edge (h5);%
      
      \draw [dashed] (17.85, -1) -- (17.85, 9);
      \node[input node] (txt1) at (15.8, 8.7) {encoder};%
      \node[input node] (txt2) at (20, 8.7) {decoder};%
      \end{tikzpicture}%
 }
 
 \newcommand{\networkTrainingPictureSmall}{%
 \begin{tikzpicture}[shorten >=1pt,->,draw=black!50, font=\footnotesize, scale=0.48]%
     \tikzstyle{node}=[circle,fill=gray!20,minimum size=25pt,inner sep=0pt]
     \tikzstyle{box}=[rectangle,draw=black!50,minimum height=12pt, inner sep=0pt]
     \tikzstyle{input node}=[draw=none, fill=none];
     \tikzstyle{network node}=[box, draw, fill=gray!20, minimum width=32pt];
     \tikzstyle{output node}=[box, minimum width=42pt,inner sep=2pt];
     \tikzstyle{projection node}=[box, minimum width=42pt,inner sep=2pt];
     \tikzstyle{spline node}=[box, minimum width=42pt,inner sep=2pt];
     \tikzstyle{sample node}=[draw=none, fill=none];
     \tikzstyle{loss node}=[box, minimum width=42pt,inner sep=2pt];
     \tikzstyle{node dots}=[node, fill=black, scale=0.2];
     \tikzstyle{annot} = [text width=4em, text centered]
 
      \node[draw=none, fill=none] (h0) at (1.5,1.8){$h_{0}$};
 
      \foreach \i/\j in {1/0, 2/1} {%
        \node[input node] (x\i) at (4*\i,0 ) {$z_{\j}, x_{\i}$};%
        \node[network node] (h\i) at (4*\i,1.8) {$h_{\i}$};%
        \node[projection node] (p\i) at (4*\i,3.6) {$\alpha_{\i} | h_{\i}$};%
        \node[loss node] (l\i) at (4*\i,5.4) {$\mathcal{L}_{\i}$};
        
        \path (x\i) edge (h\i);%
        \path (h\i) edge (p\i);%
        \path (p\i) edge (l\i);%
      }%
 
 
      \node[input node] (d1) at (11.5, 1.8) {$\dots$};%
      \path (h0) edge (h1);%
      \path (h1) edge (h2);%
      \path (h2) edge (d1);%
 
      \foreach \c/\i/\j in {3/{T}/{T-1}} {%
        \node[input node] (x\c) at (5*\c, 0) {$z_{\j}, x_{\i}$};%
        \node[network node] (h\c) at (5*\c,1.8) {$h_{\i}$};%
        \node[projection node] (p\c) at (5*\c,3.6) {$\alpha_{\i} | h_{\i}$};%
        \node[loss node] (l\c) at (5*\c,5.4) {$\mathcal{L}_{\i}$};
        
        \path (x\c) edge (h\c);%
        \path (h\c) edge (p\c);%
        \path (p\c) edge (l\c);%
      }%
      \path (d1) edge (h3);%
      \end{tikzpicture}%
 }
 
 \newcommand{\networkInferencePictureSmall}{%
 \begin{tikzpicture}[shorten >=1pt,->,draw=black!50, font=\footnotesize, scale=0.48]%
     \tikzstyle{node}=[circle,fill=gray!20,minimum size=25pt,inner sep=0pt]
     \tikzstyle{box}=[rectangle,draw=black!50,minimum height=12pt, inner sep=0pt]
     \tikzstyle{input node}=[draw=none, fill=none];
     \tikzstyle{network node}=[box, draw, fill=gray!20, minimum width=32pt];
     \tikzstyle{output node}=[box, minimum width=42pt,inner sep=2pt];
     \tikzstyle{projection node}=[box, minimum width=42pt,inner sep=2pt];
     \tikzstyle{spline node}=[box, minimum width=42pt,inner sep=2pt];
     \tikzstyle{sample node}=[draw=none, fill=none];
     \tikzstyle{loss node}=[box, minimum width=42pt,inner sep=2pt];
     \tikzstyle{node dots}=[node, fill=black, scale=0.2];
     \tikzstyle{annot} = [text width=4em, text centered]
 
      \node[draw=none, fill=none] (h0) at (1.5,1.8){$h_{T}$};
 
      \foreach \c/\i/\j in {1/{T+1}/{T}} {%
       \node[input node] (x\c) at (5*\c, 0) {$z_{\j}, x_{\i}$};%
       \node[network node] (h\c) at (5*\c,1.8) {$h_{\i}$};%
       \node[projection node] (p\c) at (5*\c,3.6) {$\alpha_{\i} | h_{\i}$};%
       \node[sample node] (l\c) at (5*\c,5.4) {$\hat{z}_{\i}$};
         
       \path (x\c) edge (h\c);%
       \path (h\c) edge (p\c);%
       \path (p\c) edge (l\c);%
       }
         
      \foreach \c/\i/\j in {2/{T+2}/{T+1}} {%
        \node[input node] (x\c) at (5*\c, 0) {$\hat{z}_{\j}, x_{\i}$};%
        \node[network node] (h\c) at (5*\c,1.8) {$h_{\i}$};%
        \node[projection node] (p\c) at (5*\c,3.6) {$\alpha_{\i} | h_{\i}$};%
        \node[sample node] (l\c) at (5*\c,5.4) {$\hat{z}_{\i}$};
        
        \path (x\c) edge (h\c);%
        \path (h\c) edge (p\c);%
        \path (p\c) edge (l\c);%
      }%
 
      \node[input node] (d2) at (13, 1.8) {$\dots$};%
      \path (h0) edge (h1);%
      \path (h1) edge (h2);%
      \path (h2) edge (d2);%
 	
      \draw [dashed] (5., 4.9) -- (8.8, 0.3);
 	
      \foreach \c/\i/\j in {3/{T + \tau}/{T+\tau-1}} {%
        \node[input node] (x\c) at (5.3*\c,0 ) {$\hat{z}_{\j}, x_{\i}$};%
        \node[network node] (h\c) at (5.3*\c,1.8) {$h_{\i}$};%
        \node[projection node] (p\c) at (5.3*\c,3.6) {$\alpha_{\i} | h_{\i}$};%
        \node[sample node] (l\c) at (5.3*\c,5.4) {$\hat{z}_{\i}$};
        
        \path (x\c) edge (h\c);%
        \path (h\c) edge (p\c);%
        \path (p\c) edge (l\c);%
      }%
      
      \path (d2) edge (h3);%
      
 	
      \end{tikzpicture}%
 }
 
 \newcommand{\networkInferencePicture}{%
 \begin{tikzpicture}[shorten >=1pt,->,draw=black!50, font=\footnotesize, scale=0.48]%
     \tikzstyle{every pin edge}=[<-,shorten <=1pt]
     \tikzstyle{node}=[circle,fill=gray!20,minimum size=25pt,inner sep=0pt]
     \tikzstyle{box}=[rectangle,draw=black!50,minimum height=12pt, inner sep=0pt]
     \tikzstyle{input node}=[draw=none, fill=none];
     \tikzstyle{network node}=[box, draw, fill=gray!20, minimum width=32pt];
     \tikzstyle{output node}=[box, minimum width=42pt,inner sep=2pt];
     \tikzstyle{projection node}=[box, minimum width=42pt,inner sep=2pt];
     \tikzstyle{spline node}=[box, minimum width=42pt,inner sep=2pt];
     \tikzstyle{loss node}=[box, minimum width=42pt,inner sep=2pt];
     \tikzstyle{sample node}=[node];
     \tikzstyle{node dots}=[node, fill=black, scale=0.2];
     \tikzstyle{annot} = [text width=4em, text centered]
 
      \node[draw=none, fill=none] (h0) at (1.5,1.8){$\mathbf{h}_{i, 0}$};
 
      \foreach \i/\j in {1/0} {%
        \node[input node] (x\i) at (4*\i,0 ) {$z_{i,\j}, \mathbf{x}_{i, \i}$};%
        \node[network node] (h\i) at (4*\i,1.8) {$\mathbf{h}_{i, \i}$};%
        \node[projection node] (p\i) at (4*\i,3.6) {$\theta_{i,\i} | \mathbf{h}_{i, \i}$};%
        \node[spline node] (s\i) at (4*\i,5.2) {$q_{\theta_{i,\i}}(\cdot$)};%
        
        \path (x\i) edge (h\i);%
        \path (h\i) edge (p\i);%
        \path (p\i) edge (s\i);%
      }%
 
      \node [left of=x1, node distance=1.05in, align=flush right] (xtlabel) {inputs};%
      \node [left of=h1, node distance=1.0in, align=flush right] (htlabel) {network};%
      \node [left of=p1, node distance=0.95in, align=flush right] (ptlabel) {projection};%
      \node [left of=s1, node distance=1.06in, align=flush right] (stlabel) {spline};%
      \node [left of=l1, node distance=1.03in, align=flush right] (ltlabel) {sample};%
 
      \node[input node] (d1) at (7, 1.8) {$\dots$};%
      \path (h0) edge (h1);%
      \path (h1) edge (d1);%
 
     \foreach \c/\i/\j in {2/{T_i}/{T_i-1}} {%
       \node[input node] (x\c) at (5*\c, 0) {$z_{i,\j}, \mathbf{x}_{i, \i}$};%
        \node[network node] (h\c) at (5*\c,1.8) {$\mathbf{h}_{i, \i}$};%
        \node[projection node] (p\c) at (5*\c,3.6) {$\theta_{i,\i} | \mathbf{h}_{i, \i}$};%
        \node[spline node] (s\c) at (5*\c,5.2) {$q_{\theta_{i,\i}}(\cdot$)};%
              
        \path (x\c) edge (h\c);%
        \path (h\c) edge (p\c);%
        \path (p\c) edge (s\c);%
        }
        
      \foreach \c/\i/\j in {3/{T_i+1}/{T_i}} {%
       \node[input node] (x\c) at (5*\c, 0) {$z_{i,\j}, \mathbf{x}_{i, \i}$};%
       \node[network node] (h\c) at (5*\c,1.8) {$\mathbf{h}_{i, \i}$};%
       \node[projection node] (p\c) at (5*\c,3.6) {$\theta_{i,\i} | \mathbf{h}_{i, \i}$};%
       \node[spline node] (s\c) at (5*\c,5.2) {$q_{\theta_{i,\i}}(\cdot$)};%
       \node[sample node] (l\c) at (5*\c,7) {$\hat{z}_{i, \i}$};
         
       \path (x\c) edge (h\c);%
       \path (h\c) edge (p\c);%
       \path (p\c) edge (s\c);%
       \path (s\c) edge (l\c);%
       }
         
      \foreach \c/\i/\j in {4/{T_i+2}/{T_i+1}} {%
        \node[input node] (x\c) at (5*\c, 0) {$\hat{z}_{i,\j}, \mathbf{x}_{i, \i}$};%
        \node[network node] (h\c) at (5*\c,1.8) {$\mathbf{h}_{i, \i} | \Theta$};%
        \node[projection node] (p\c) at (5*\c,3.6) {$\theta_{i,\i} | \mathbf{h}_{i, \i}$};%
        \node[spline node] (s\c) at (5*\c,5.2) {$q_{\theta_{i,\i}}(\cdot$)};%
        \node[sample node] (l\c) at (5*\c,7) {$\hat{z}_{i, \i}$};
        
        \path (x\c) edge (h\c);%
        \path (h\c) edge (p\c);%
        \path (p\c) edge (s\c);%
        \path (s\c) edge (l\c);%
      }%
 
      \node[input node] (d2) at (23.5, 1.8) {$\dots$};%
      \path (d1) edge (h2);%
      \path (h2) edge (h3);%
      \path (h3) edge (h4);%
      \path (h4) edge (d2);%
 	
      \draw [dashed] (15., 6.6) -- (18.8, 0.3);
 	
      \foreach \c/\i/\j in {5/{T_i + \tau}/{T_i+\tau-1}} {%
        \node[input node] (x\c) at (5.3*\c,0 ) {$\hat{z}_{i,\j}, \mathbf{x}_{i, \i}$};%
        \node[network node] (h\c) at (5.3*\c,1.8) {$\mathbf{h}_{i, \i}$};%
        \node[projection node] (p\c) at (5.3*\c,3.6) {$\theta_{i,\i} | \mathbf{h}_{i, \i}$};%
        \node[spline node] (s\c) at (5.3*\c,5.2) {$q_{\theta_{i,\i}}(\cdot$)};%
        \node[sample node] (l\c) at (5.3*\c,7) {$\hat{z}_{i, \i}$};
        
        \path (x\c) edge (h\c);%
        \path (h\c) edge (p\c);%
        \path (p\c) edge (s\c);%
        \path (s\c) edge (l\c);%
      }%
      
      \path (d2) edge (h5);%
      
      \draw [draw=black] (10.5, -1) -- (10.5, 9);
      \node[input node] (txt1) at (10.8, 8.7) {encoder};%
      \node[input node] (txt2) at (13.65, 8.7) {decoder};%
      \end{tikzpicture}%
 }

\section{Introduction}

In large-scale distributed systems or cloud environments, the detection of anomalous events allows operators to detect and understand operational issues and facilitates swift troubleshooting. Undetected anomalies can result in potentially significant losses and can impact customers of these systems and services negatively. Designing an effective anomaly detection system is therefore an important task. This task entails significant challenges, beginning with the fact that the notion of ``anomaly'' is itself ambiguous and is used to denote different kinds of events depending on the application domain. For example, what is considered an anomaly when monitoring an EEG signal in a health care application differs from what is considered anomalous in financial time series. 
To reduce this ambiguity, we focus on anomaly detection in the context of our target application of monitoring compute systems and cloud resources, where main object of interest are metrics emitted by these systems; we refer to this setting as \emph{cloud monitoring}. We refer the reader to detailed overviews~\cite{anom1, anom2, anom3, anom4, anom5} on other application areas for anomaly detecion.

\begin{figure}[tb]
    \centering
        \includegraphics[width=.4\textwidth]{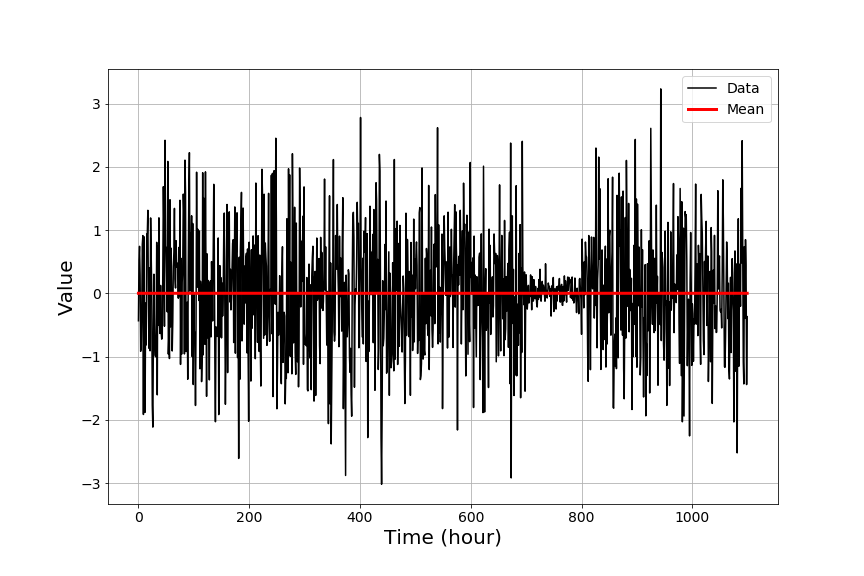}
    \caption{Synthetic example of of a collective anomaly in a time series that can only be detected by considering multiple observations simultaneously. The mean of the time series (shown in red) is constant, but for time steps 700--800 the variance of the signal descreases. Individually, each observation in this range falls into a high-density region under the distribution of previous observations and does not appear anomalous; however, collectively, these observations are clearly different from the preceding observations, following a different distribution.}\label{plot:simulated_ts}
\end{figure}

In the setting of cloud monitoring, an anomaly detection system needs to be able to efficiently detect anomalous events in a streaming fashion. The fundamental difficulties that any anomaly detection system has to face are threefold. First, due to the number and diversity of the monitored metrics (often millions) and the streaming nature of the data, it is uncommon to have sufficient amounts of labeled data available to employ supervised learning techniques. Even if labels are available, due to the subjectivity of the task, labels may not represent an ``objective'' ground truth, and there may be significant disagreement between multiple labelers. This raises the need for \emph{unsupervised} techniques. Second, the monitoring systems have to track the evolution of a large number of time series simultaneously, which often leads to a considerable flow of data to process in near real-time, so the models have to \emph{scale efficiently} to the amount of data available. Here, scalability comes not only in the traditional flavor of computational scalability, but also in terms of the need to involve experts to tune the systems. With millions of metrics to be monitored, methods are required that have robust out-of-the-box performance requiring no manual intervention. Finally, the methods have to be \emph{flexible} in order to handle time series of different nature (e.g.\ CPU usage, latency, error rate), and anomalies presenting a wide range of patterns (point anomalies, collective anomalies%
\footnote{A collective anomaly consists of a subset of points that deviates from the rest of the dataset even though individually each point may appear normal (see for example Figure \ref{plot:simulated_ts}).}%
, contextual anomalies, abrupt changes in trend).

\begin{figure*}[t]
    \centering
        \includegraphics[width=.32\linewidth]{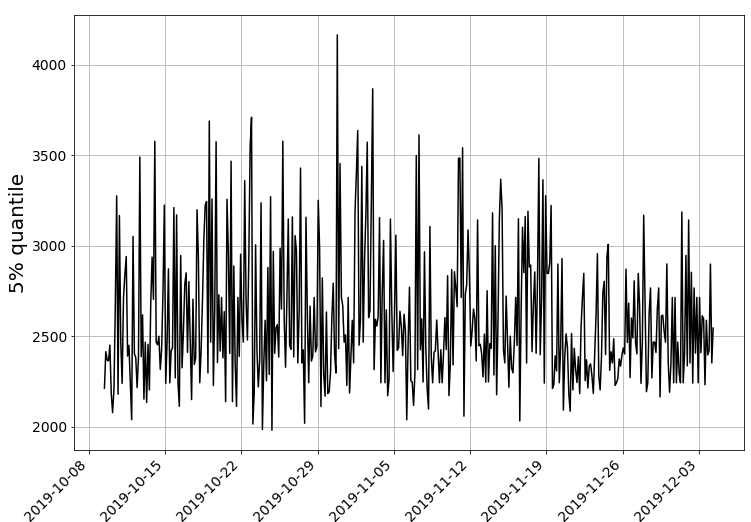}
        \includegraphics[width=.32\linewidth]{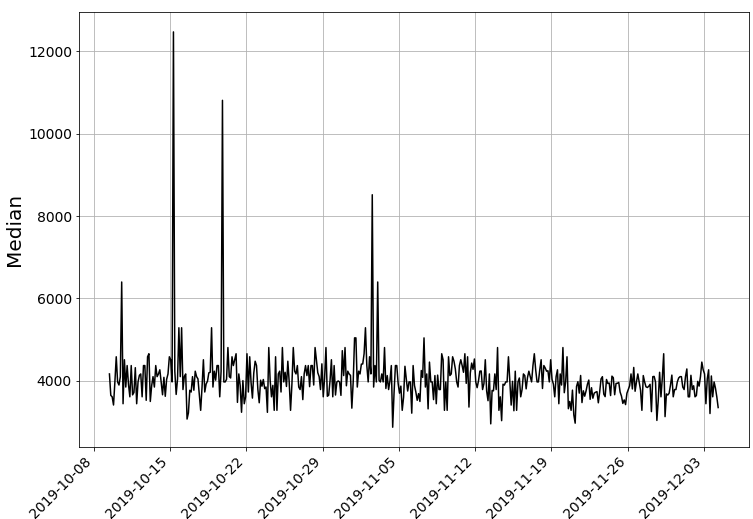} 
        \includegraphics[width=.32\linewidth]{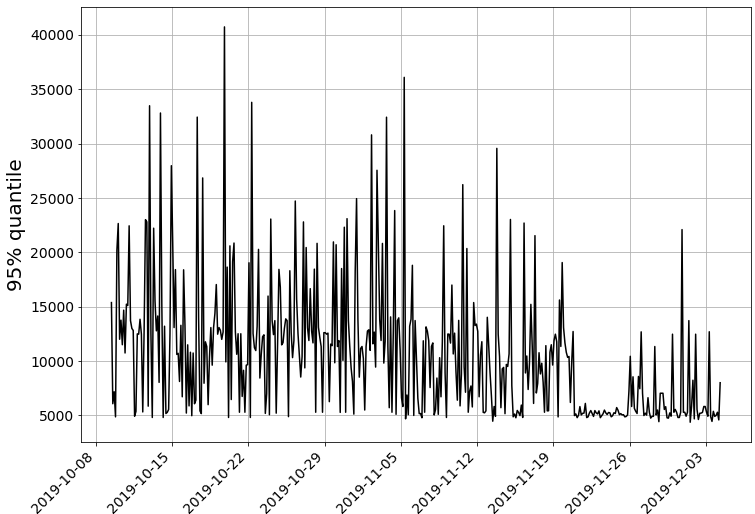}
    \caption{Latency metric monitoring with temporal aggregation using different summary statistics. The three panels show the same underlying event data (latency measurements) from an internal service, aggregated into five-minute intervals using three different summary statistics: (left) $5\%$ quantile;  (center) median; (right) $95 \%$ quantile. The anomalous region occuring at the end of November is clearly visible in the $95\%$ quantile, but harder (or impossible) to detect in the other two.}\label{plots:time_ds}
\end{figure*}

The main contribution of the present work is a novel anomaly detection method based on distributional time series models that addresses all three challenges. To the best of our knowledge it is the first anomaly detection methodology that builds on a predictive model for a distributional time series representation. It employs an autoregressive LSTM-based recurrent neural network to provide flexibility while still being statistically sound. Our model scales well at inference time and 
has a compact model state making it deployable in low-latency, streaming application scenarios. It readily allows the incorporation of covariates, enabling the model to detect contextual anomalies where the context is not limited to the temporal one. For example, a high CPU utilization may be expected if the number of incoming requests is large, but abnormal when the number of requests is low. Finally, our methodology can detect collective anomalies, which most non-distributional techniques are unable to detect. 

We evaluate our method on a number of data sets including synthetic, publicly available, and AWS-internal data sets, and show that our method compares favorably to the state of the art. While we develop our methodology for the cloud monitoring setting, we further show that our method is competitive in classical anomaly detection settings.

We proceed by first discussing a motivating example for our method and provide background in Section~\ref{sec:motivation}, 
introduce the model formally in Section~\ref{sec:model}, evaluate it empirically in Section~\ref{sec:experiments}, discuss related work in Section~\ref{sec:related_work}, before concluding in Section~\ref{sec:conclusion}.

\begin{figure*}[htb]
\centering
    \includegraphics[width=0.8\textwidth]{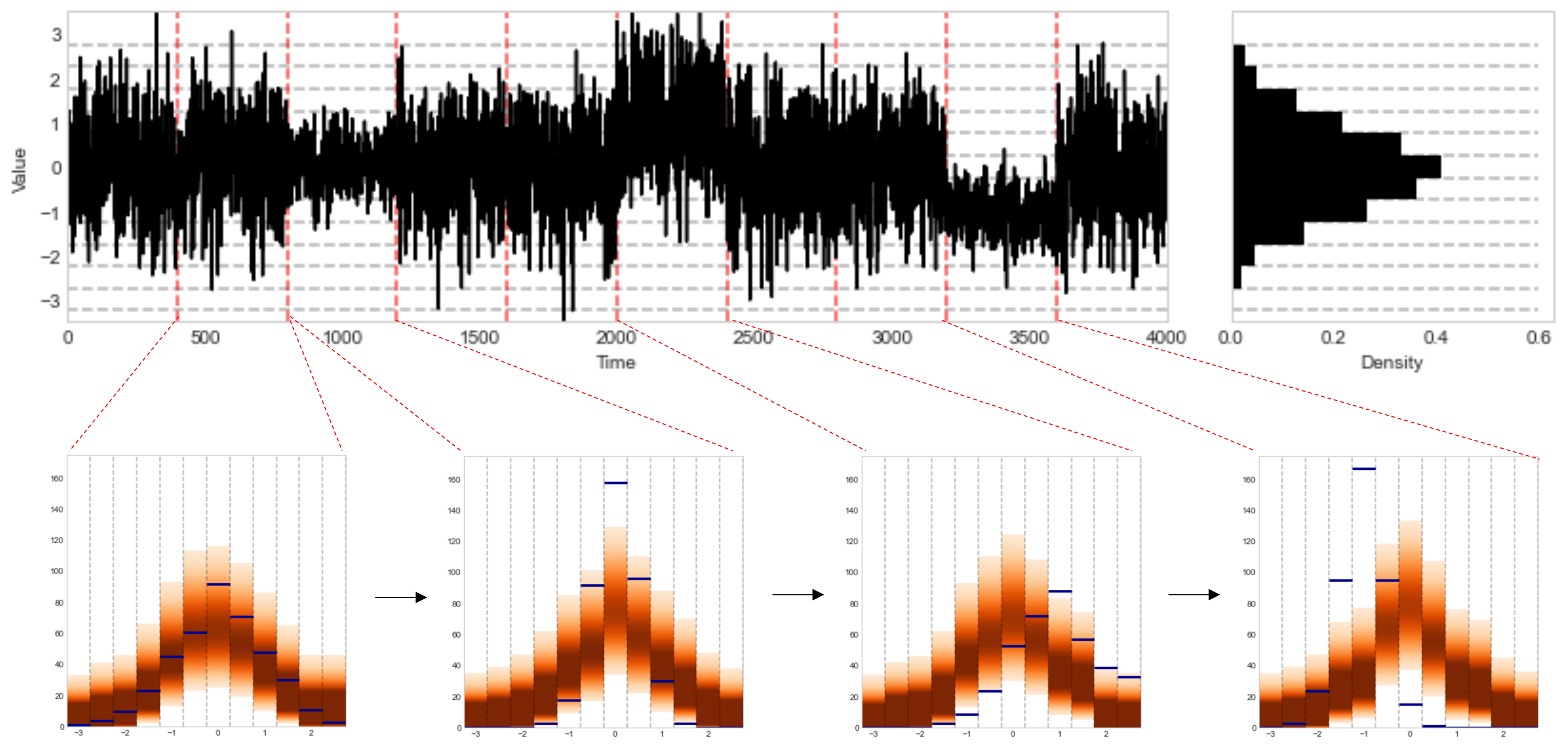}
    \caption{Illustration of our approach. The undelying signal (top panel) is grouped into fixed-size time intervals (vertical red dashed lines) of size $n_t$ (here $n_t=400$). In each interval, we estimate the probability distribution $\tilde{F}_t$ of the values within the interval using a histogram (blue horizontal bars in the bottom panels), with bin edges (dashed grey lines) chosen according to a global strategy (e.g.\ based the marginal distribution, top right panel). For each time interval, the model predicts a probability distribution over probability distributions (illustrated by the yellow-red heatmaps in the bottom panel) based on the preceding intervals and a time-evolving hidden state $h_t$ using a RNN. For ``normal'' periods (e.g. bottom left panel), the observed data (blue lines) aligns with the model's prediction, i.e.\ the blue lines fall into the shaded area. For ``anomalous'' periods (three rightmost bottom panels), the model's prediction differs from the observed data, i.e.\ the observed histogram falls outside the high-probability region predicted by the model.
    } \label{fig:overview}
\end{figure*}

\section{Motivation}\label{sec:motivation}
In the following we motivate our \emph{distributional} time series modeling approach from two angles: the data generation process of request-driven metrics, and high-frequency time series. 

In a typical (micro-) service monitoring setup, a metric datum is emitted for each request handled by the service, containing information about e.g.\ the processing time for the request and whether it resulted in a success or failure (and potentially more fine-grained information about the request and the response). The raw monitoring data is thus a stream of events, where each event is a tuple consisting of a timestamp and a set of measurements.
As a measurement is triggered for each incoming request, the time stamps are not equally spaced, and for large services one may collect hundreds of thousands of events per minute.
To facilitate further processing and modeling, the typical anomaly detection pipeline starts with a temporal aggregation step, where the event data is aggregated into fixed-sized time intervals (e.g.\ one minute), recovering the classical, equally-spaced time series setting. This aggregation of events requires choosing a meaningful statistic which summarizes all measurements within a given time interval, while allowing detection of abnormal behaviors. Commonly used summary statistics are e.g.\ the mean (or median), and extreme percentiles (e.g.\ the maximum, minimum, or the 99th percentile). However, the summary statistics chosen ultimately determine the range of anomalies one can detect, and one risks missing anomalies if the statistics are chosen inappropriately. For example, in Figure \ref{plots:time_ds}, we show three different quantiles of latency measurements of an internal service handling a large number of requests. We observe that if we choose to monitor the median or the 5\% quantile, we would miss the anomaly that can be seen in the 95\% quantile. 

The method we propose here embraces this event-based data generation process by considering the entire \emph{distribution} of measurements within each time interval. This means considering time series of equally spaced ``points'' in time, but where each ``point'' is a probability distribution, called a \emph{distributional data point}. This is in contrast to most classical anomaly detection approaches that take a time series of equally-spaced, real-valued data points as input and do not explicitly model the temporal data aggregation step.

Even though the proposed method was originally designed for the particular nature of the data described above, we demonstrate highly competitive performance even in the ``classical'' setting, where the starting point are time series of real values sampled at a regular frequency. We discuss 
this via the example of high-frequency time series, arising for example from measuring the CPU utilization or temperature of a compute node every second. Our approach solves several difficulties specific to such metrics. 

The main challenge one faces when performing anomaly detection for high-frequency data is that the temporal dynamics governing the data evolve at a slower pace than the frequency of observation. In typical application settings, meaningful variations in metrics are expected to occur from one hour to the next, but not every second. The underlying dynamics can thus often be adequately described by using one hour or half an hour time granularity, with seasonal patterns that are daily, weekly or even monthly. However, both classical and deep-learning-based time series models are commonly unable to model long range dependencies (measured in number of observations), so that if high-frequency data is modeled directly, these models commonly fail to capture medium and long term patterns.

Our approach allows modeling the the temporal evolution at a more appropriate frequency by aggregating the observations, while retaining the ability to detect anomalies at the original frequency by modeling the distribution of observations within each time interval. 
Within each aggregated time interval $t$, we treat the high-frequency data point as samples from this distribution. As an illustration, consider time series with a one-minute sampling frequency. If aggregated hourly, this yields $n_t = 60$ observations per aggregated time interval. Based on the preceding one-hour data distributions, our approach predicts the distribution of the observations for the hour to come. Then, at the $m$-th minute of that hour, we compute the likelihood of the current observation, which is used to determine if it is anomalous or not. We can also compute the joint likelihood of the past $m$ observations in the hour, allowing us to detect collective anomalies. An example of such anomalies is given in Figure \ref{plot:simulated_ts}. We can see that the variance of the data distribution decreases drastically: individually, each observation falls into a high-density region under the distribution of recent observations and does not appear as an outlier; however, observing these $m$ values in a row is highly improbable. Classical time series anomaly detection algorithms are not able to detect such anomalies.

\section{Model}\label{sec:model}

The backbone of our anomaly detection technique is a deep probabilistic distributional time series model, i.e.\ a probabilistic model for time series $x_{1:T} = x_1, x_2, \ldots, x_T$, where each element $x_t$ is a probability distribution over real values. See Figure \ref{fig:overview} for an illustration. This is in contrast to the traditional setting where the time series elements themselves are real values. In the following, we introduce the necessary notation and tools used in the rest of the paper.
We start by briefly giving a high-level description of the generic anomaly detection methodology with a probabilistic model. More details can be found for example in~\cite{Faloutsos2018}. \\

\indent \textbf{Anomaly detection with probabilistic models.}
One generic strategy for anomaly detection using probabilistic models is to compute the predictive distribution under the model, and mark an observation as anomalous if its probability under this distribution is low.
In the time series setting, i.e.\ where the observations $x_{1:T}$ are a generic time series taking values $x_t \in \mathcal{X}$, a probabilistic time series model aims at modeling $\mathbb{P}_{T+1}(\cdot \ | x_{1:T})$, i.e.\ the conditional distribution of the next value $x_{T+1}$ given the history of observations $x_{1:T}$.  This conditional model can take various forms, the most commonly used one being a parametric model whose parameters are determined by the preceding observations. Once we construct an estimate $\widehat{\mathbb{P}}_{T+1} (\cdot \ | x_{1:T})$, we can use it to define a credible region and mark any observation $x_{T+1}$ that falls outside of this credible region as anomalous. 

In the classical time series anomaly detection setting, the observation space $\mathcal{X}$ is taken to be $\mathbb{R}^D$ ($D \geq 1$), whereas here we take it to be the space of probability distributions on $\mathbb{R}$. 

\subsection{Distributional Time Series}
Let $F_{1:T} = F_1, F_2, \ldots, F_T$ be a time series of univariate probability distributions, represented by their cumulative distribution functions (CDFs). 
We assume that the support for all $F_t$ is the interval $\mathbb{Y} = [ y_{\min}, y_{\max}]$.%
\footnote{In practice, we need to determine these bounds for each monitored time series. One strategy is to choose them as a function of extreme values observed historically.}

Even though these distributions are the objects of interest, we usually do not to have access to them directly. Because of this, we also consider the scenario where we observe $F_t$ only indirectly through samples, i.e.\ at each time $t$ a set $\mathcal{Y}_t = \{ Y_{t1},..., Y_{tn_t}\}$ of $n_t$ iid samples from $F_t$ is observed. Depending on the value of $n_t$, we can differentiate three real-world use cases:
\begin{enumerate}
\item \textbf{Monitoring services with frequent requests:} This corresponds to the setting described in Section~\ref{sec:motivation}, where for each time interval (e.g.\ each minute), the number of measurements $n_t$ is large, e.g.\ on the order of $10^5$ or more. The underlying distributions $F_t$ can then be estimated with a high enough precision for us to consider that they are directly observed. We will also refer to this as the \emph{asymptotic settting}, since it corresponds to our model when $n_t \rightarrow +\infty$. 
\item \textbf{High-frequency time series:} This corresponds to the setting where the temporal resolution of the original time series is higher than the scale at which meaningful temporal variation occurs. As modeling long-range dependencies is challenging for both classical and deep-learning-based time series models, it can be desirable to aggregate the observations to a lower, more meaningful time granularity. This leads to the case where $n_t > 1$ but small, e.g.\ $n_t=60$ when aggregating from seconds to minutes. In this setting $n_t$ is usually constant, as the underlying time series will typically have a regular time-granularity.
\item \textbf{Low-frequency time series:} We also consider the $n_t = 1$ setting, where our model reduces to a classical probabilistic time series model over real-values observations. Even though this is not the setting for which our approach was originally designed, we will show in the experiment section that it still yields competitive results.
\end{enumerate}
Our model handles all three settings. We will refer to the last two scenarios as the \emph{finite $n_t$ scenarios}, in contrast to the first one. In the asymptotic setting, the distributions $F_t$ are observed directly. In the finite $n_t$ settings, $F_{1:T}$ are unobserved and we only observe samples from them. Therefore, we need to be able to assess the likelihood of $F_t$ for the asymptotic regime, and the likelihood of $\mathcal{Y}_t$, where $F_t$ is marginalized out, for the finite $n_t$ regimes.

\subsection{Probabilistic Model on Binned Densities}
A common approach to modeling distributional data is to represent the functions of interest (e.g.\ the CDFs or PDFs) by a point in a carefully chosen finite-dimensional space. In this work, we will consider the space of piece-wise linear functions to approximate the CDFs, or equivalently, the space of binned (piecewise-constant) distributions to approximate the PDFs.%
\footnote{Other choices exist, which make different trade-offs: In the field of functional data analysis, where the objects of interest are general functions, not probability distributions, a common choice for the finite-dimensional representation are coordinates with respect to a truncated orthonormal basis of functions, composed of e.g.\ sinusoidal, wavelet or functional principal components. These function bases are not natural when modeling probability distributions, as it is non-trivial to enforce the required constraints: non-negativity and monotonicity for CDFs, non-negativity and unit integral for PDFs. Previous approaches employing such a representation usually resort to a post-processing step that enforces these constraints. Another finite-dimensional representation is common in the probabilistic modeling literature, where densities are often modeled as a finite mixture of base densities, typicallly chosen to be Gaussian. This representation is by construction suitable for distributional data, however obtaining such a decomposition is computationally expensive, even for a single given $F_t$.}
Specifically, we chose to approximate each CDF $F_t$ by a piece-wise linear function $\widetilde F_t$, composed of $d$ linear pieces. A given function in this class is specified by two sets of parameters: the start and end points of linear pieces (the \emph{knot positions}), and the slopes in each segment. While it is possible to adapt the knot positions dynamically (as done in \cite{gasthaus2019probabilistic}), we keep the knot positions fixed and only model the temporal evolution of the slopes within each segment. 

We divide $\mathbb{Y}$ into $d$ bins using the grid $y_{\min} = a_0 < a_1 < ... < a_{d} = y_{\max}.$ Let $\widetilde F_t$ be the piece-wise linear CDF that interpolates the points $(a_k, F_t(a_k))_{k=0,...,d}$. Therefore, the corresponding density function $\widetilde f_t$ is piece-wise constant, and the probability of falling into one of the bins $[a_{k-1}, a_k)$ is given by
\begin{equation}\label{def_pt}
    p_{tk} = F_t(a_k)-F_t(a_{k-1}).   
\end{equation}
Specifying a distribution on the $d$ dimensional probability vector $p_t = (p_{t1},...,p_{td})$ entails a distribution over the piece-wise linear CDFs $\widetilde F_t$.
We model this distribution over probability vectors using a Dirichlet distribution, i.e.\ $p_t \sim \text{Dir}(\alpha_t)$, 
$$\text{Dir} (p; \alpha) = \frac{\Gamma\left(\sum_{i=1}^d \alpha_i\right)}{\prod_{i=1}^d \Gamma(\alpha_i)} \prod_{i=1}^d p_i^{\alpha_i - 1},$$
where $\alpha_t \in \mathbb{R}_+^d$ denotes the concentration parameter whose temporal evolution is modeled using an RNN.

Before proceeding let us emphasize the fact that one can approximate any $F_t$ arbitrarily well as the grid becomes finer ($d$ becomes larger). Indeed, for any random variable $Y_t$ with distribution $F_t$, and $\widetilde Y_t$ with distribution $\widetilde F_t$ (its piecewise linear approximation as defined above), as $\max_k |a_k-a_{k-1}| \rightarrow 0$, $\widetilde Y_t$ will converge in distribution to $Y_t$. 
Based on the previous remark, we will simplify the setting and drop the notation $\widetilde F_t$, and  assume that the $F_t$ themselves are piece-wise linear. To simplify notation further (and without loss of generality), we will take $\mathbb{Y} = [0,1]$.

Given $F_t$ as described above, we have that the probability of a single observation $Y_{t1}$ given $F_t$ is $p_{tk}$ if it falls in the interval $[a_{k-1}, a_k)$, 
$$ \mathcal{L}(Y_{t1}; F_t) = \text{Cat}\big( (\mathbb{I}\{Y_{t1} \in [a_{k-1}, a_k)\})_{k=1,..,d}; p_t \big),$$
where $\mathbb{I}\{A\}$ denotes the indicator of the event $A$, and $\text{Cat}(\cdot; p_t)$ refers to the categorical distribution with parameter $p_t$. Hence, the count vector $m_t = (m_{t1}, ..., m_{td})$ with $m_{tk} = \sum\limits_{i=1}^{n_t} \mathbb{I}\{Y_{ti} \in [a_{k-1}, a_k)\}$, is a sufficient statistic, and the likelihood for a set of observations $\mathcal{Y}_t$ is given by
\begin{equation}
    \mathcal{L}(\mathcal{Y}_t; F_t) = \text{Mult}(m_t; n_t, p_t),
\end{equation}
where $\text{Mult}(\cdot; n_t, p_t)$ refers to the Multinomial distribution with $n_t$ trials and outcome probabilities $p_t$. Since we take $p_t$ to be Dirichlet-distributed, it can be marginalized out and we have a closed form probability mass function for the observations $m_t$. More precisely, $m_t$ follows a Dirichlet-Multinomial distribution with $n_t$ number of trials and concentration vector $\alpha_t$, with probability mass function, 
$$ \text{Dir-Mult} (m; n, \alpha) = \frac{\left(n!\right)\Gamma\left(\alpha_0\right)}
{\Gamma\left(n+\alpha_0\right)}\prod_{i=1}^d\frac{\Gamma(m_{i}+\alpha_{i})}{\left(m_{i}!\right)\Gamma(\alpha_{i})}. $$

To summarize, given $\alpha_t$, the likelihood of the observation is:
\begin{align*}
    \mathcal{L}_t = \mathcal{L}(p_t; \alpha_t) &= \text{Dir}(p_t; \alpha_t)  \tag{Asymptotic setting} \\
    \mathcal{L}_t = \mathcal{L}(m_t; n_t, \alpha_t) &= \text{Dir-Mult}(m_t; n_t, \alpha_t)  \tag{Finite $n_t$ setting},
\end{align*}
where as explained previously in the asymptotic regime we suppose that we directly observe $p_t$ which is equal to the normalized counts $\frac{1}{n_t} m_t$.

\subsection{RNN Temporal Dynamics Model}
In both settings, the temporal evolution of the data is described through the time-varying parameter $\alpha_t$,
and it is this dynamic behavior that we aim to learn. In order to do so, we will use an autoregressive LSTM-based recurrent neural network, whose architecture follows the one described in \cite{salinas2019deepar}. Figures \ref{fig:training} and \ref{fig:prediction} in the appendix illustrate the model's architecture for the training and prediction steps. 

Recurrent neural networks (RNNs) form a class of artificial neural networks designed to handle sequential data. They have been successfully used in a wide range of applications such as time series, natural language processing and speech recognition. One of the key benefits of RNNs is their ability to handle sequences of varying lengths. 
RNNs sequentially update a hidden state $h$: at every time step $t$, the next hidden state $h_t$ is computed by using the previous $h_{t-1}$ and the next input (in a time series this can be the next observation $y_t$ and other covariates). A crucial detail is that the weights of the network are shared across time steps, which makes the RNN \textit{recurrent}, and capable of handling sequences of varying length. The hidden states can be seen as a dynamic memory of a feature representation of the raw input. This compact representation makes them amenable to streaming settings. Here, we mainly rely on long short-term memory networks (LSTM), the arguably most popular subclass of RNNs. 

Let $z_{1:T}$ be the sequence of observations, either $p_{1:T}$ or $m_{1:T}$ depending on the setting. Denote $\phi$ the parameters of the RNN model. Given a horizon $\tau$, the aim is to predict the probability distribution of future trajectories $z_{T+1:T+\tau}$, with the potential use of observed covariates $x_{1:T+\tau}$ . 

The parameter $\alpha_t$ is function of the output $h_t$ of an autoregressive recurrent neural network with
\begin{eqnarray}
h_{t} &=& r_\phi(h_{t-1}, z_{t-1}, x_{t}) \\
\alpha_t &=& \theta_\phi(h_t)
\end{eqnarray}
where $r_\phi$ is a multi-layer recurrent neural network with LSTM cells.
The model is autoregressive and reccurent in the sense that it uses respectively the observation at the last time step $z_{t-1}$ and the previous hidden state $h_{t-1}$ as input.
Then a layer $\theta_\phi$ projects the output $h_t$ to $\mathbb{R}_+^d$, the domain of $\alpha_t$. The parameters $\phi$ of the model are chosen to minimize the negative log likelihood:
$$ \bm{\text{L}} = - \sum\limits_{t=1}^T \log (\mathcal{L}_t).$$
Finally, we note that when dealing with anomaly detection we only require a time horizon $\tau = 1$.

\subsection{Anomaly Detection with Level Sets}
Once we forecast $\alpha_{T+1}$, we can assess whether the observation $z_{T+1}$ is a potential anomaly. Indeed, given $\alpha_{T+1}$, we know the distribution of the random variable $Z_{T+1}$, of which $z_{T+1}$ should be a sample if no anomaly happened. We can consequently compute a credible region $\mathcal{C}_{T+1}$ with total mass $1-\varepsilon$ for a given level $\varepsilon$. Then, if $z_{T+1} \not \in \mathcal{C}_{T+1}$, we will say that the observation is an anomaly. The difficulty one faces when considering credible regions is that they are not unique. Even though this problem exists for an univariate setting, it can be easily circumvented and natural credible intervals can be designed. In a multivariate setting, this issue is more challenging and one needs to choose meaningful credible regions. 
The credible regions we will consider are the level-sets of the likelihood, defined by 
$$ \mathcal{S}_{T+1}(\eta) = \{ z : \mathcal{L}_{T+1}(z) \geq \eta )\}. $$
We will then take $\eta_{T+1}$ such that 
$$\mathbb{P} (Z_{T+1} \in \mathcal{S}_{T+1}(\eta_{T+1})) = 1- \varepsilon,$$
and $ \mathcal{C}_{T+1} = \mathcal{S}_{T+1}(\eta_{T+1})$. In other words, the credible region will be the highest density region that achieves a total mass of $1-\varepsilon$, and the observation will be considered as an anomaly if $\mathcal{L}_{T+1}(z_{T+1}) < \eta_{T+1}$. The remaining difficulty is to compute $\eta_{T+1}$. This theoretically requires computing the mass of the level-sets, and then invert the function $\eta \mapsto \mathbb{P} (Z_{T+1} \in \mathcal{S}_{T+1}(\eta))$. When the number of possible outcomes for $Z_{T+1}$ is finite and relatively small, this can be done exactly by computing the likelihoods of all outcomes. Otherwise, we will approximate such inverse function by means of a Monte Carlo method, following an idea that goes back to \cite{hyndman1996computing}. If we consider the univariate random variable defined as $\mathcal{L}_{T+1}(Z_{T+1})$, we remark that $\eta_{T+1}$ can be interpreted as the $\varepsilon$ quantile of that distribution. Therefore, we will construct the following estimator $\hat \eta_{T+1}$: first we sample $M$ realizations of $Z_{T+1}$, then we compute the associated $M$ likelihoods, and finally take $\hat \eta_{T+1}$ the $\varepsilon$ quantile of their empirical distribution.   

For the asymptotic or the low frequency setting, we simply apply the approach described above. For the high frequency setting, we use a two-stage approach. For the sake of exposition we will describe the procedure on the following illustrative example. Suppose that we observe a minute frequency time series, and suppose we are interested in hourly aggregation. Suppose that we have  $T$ hours of observations, which when aggregated give $\mathcal{Y}_{1:T}$ sets of observations, where $\mathcal{Y}_t = \{ Y_{t, 1}, ..., Y_{t, 60}  \}$. From the forecasting module, we predict $\alpha_{T+1}$ and hence the distribution of the observations for the hour to come. Every minute we obtain a new observation. In the first stage, before the hour is over, we assess every minute whether the current observation is anomalous. For this stage, there are only $d$ possible outcomes, since the observation will fall in one of $d$ possible bins, therefore we can compute the level sets exactly without Monte Carlo estimation. Once the hour is over, we assess whether the past $60$ observations jointly constitute a collective anomaly. This is done by constructing the count vector $m_{T+1}$ and checking if it falls within the credible region. Here, since the number of possible outcomes is too large, we will use a Monte-Carlo estimate. If we want to detect collective anomalies that are shorter, we can add an intermediate stage.

For example every $15$ minutes we can assess whether the previous $15$ observations are jointly anomalous.
Finally, as explained in the experiment section, we will need to give an anomaly score to each time point to evaluate the models. The score used is the logarithm of the p-value, which is the smallest $\epsilon$ for which a given point is considered as an anomaly. For the two-stage strategy, we simply add the two scores.

\section{Experiments}\label{sec:experiments}

Our implementation\footnote{The code is available at \url{https://github.com/awslabs/gluon-ts/tree/distribution_anomaly_detection/distribution_anomaly_detection}.} is based on GluonTS~\cite{alexandrov2020gluonts} which in turn is based on MXNet~\cite{chen2015mxnet}. 
Since we inherit scalability for both training and inference from MXNet, we do not explicitly 
perform experiments with respect to computational efficiency and scalability and resort to report the following details. Note that we learn a global model (across all metrics) which takes roughly 3mins per 100 metrics. For such models, we do not have to re-train often, so we may disregard the training time for 
the production scenario. Inference scales embarrasingly parallel. Scoring of a single data point take 1ms for 1 minutely aggregated data (note that we do not perform the costly Monte-Carlo estimates at every time point). We can limit memory consumption of the models to a fixed size of 80kb per metric. We note, that our model runs in a streaming setting where data is provided by Amazon Kinesis. We can process approximately 65k metrics per minute on a standard 16-core EC2 instance and scale out horizontally to millions of metrics 
effortlessly.

For all the experiments we learn the parameters of the model on the learning time range $\{ 0,...,T \}$, and we perform anomaly detection on the detection time range $\{ T+1, ..., T+ D\}$. We fix the prediction time length to $\tau=1$, and perform anomaly detection in a streaming way:  observing $z_{1:T}$, we predict the distribution of $Z_{T+1}$, then assess the likelihood of the next observation $z_{T+1}$. Then, knowing $z_{1:T+1}$, we predict the distribution of $Z_{T+2}$ and compare it to the next observation $z_{T+2}$ and so on. We consider two different grids to define the bins. The first one is the simple regular grid, $a_k = k/d$. The second grid is obtained using the quantiles of the marginal distribution; in the asymptotic case we compute the average of the observed cdfs, in the finite $n_t$ setting we concatenate all the sets of observations; then for both cases we take the $d+1$ quantiles with regularly spaced levels. Depending on the problem, the regular or the quantiles grid can be better. Indeed, the quantiles grid can give poor results if some distributions are very different from the marginal, since this approach can leave large regions with very few bins. Another approach could be to consider a time varying grid, for example consider the quantiles of the marginal distribution of each day of the week, but we didn't experiment this approach yet.

\subsection{Evaluation metric}
For comparing the different models we will use the area under the receiver operating characteristic curve (ROC-AUC). It is a metric commonly used for classification problems to compare algorithms which performances depend on selecting a threshold. This measure quantifies how much a model is able to distinguish between the two classes. It takes values between $0$ and $1$, the higher the better. This score is independent of the threshold chosen since it only considers the ranking of the observations by the model in terms of how much abnormal it looks. Therefore it allows to quantify the maximum potential of a method.

\subsection{Synthetic data}
Let $\mu_t = \sin(\frac{2\pi t}{P})$ and $\sigma_t = 1$, where $P=24$ is a period length and $\epsilon_t \sim \mathcal{N}(0,0.1)$ are iid noise. We will consider the two following dynamics:
\begin{enumerate}
    \item\label{noise_mean} \textbf{DS1:} $F_t = \mathcal{N}(\mu_t + \epsilon_t, \sigma_t)$
    \item\label{noise_var}  \textbf{DS2:} $F_t = \mathcal{N}(\mu_t, \sigma_t + \epsilon_t)$
\end{enumerate}
We consider $T=1500$ learning time points (which corresponds to approximately 2 months of hourly data) and a detection time horizon of $D=2000$. In the detection time range, we add an anomaly with probability $3\%$ at each location independently. For each experiment, we use one of two types of anomalies: a sudden distributional shift (by adding 1 to $\mu_t$), or a distributional collapse (removing $1/2$ to the standard deviation $\sigma_t$, as in Figure \ref{plot:simulated_ts} in the introduction). We therefore get four different settings, we will denote them respectively DS1 $\mu$, DS1 $\sigma$, DS2 $\mu$ and DS2 $\sigma$.

We set the threshold for anomaly detection to be $95\%$. For the Monte Carlo approximation of the level set, we take $M=1000$ samples from the predictive distribution of the log likelihoods and estimate the corresponding $\eta$.
An observation can then be considered anomalous in two cases. In the first case, the generated noise term $\epsilon_t$ falls outside of a $95\%$ confidence interval of the $\mathcal{N}(0,0.1)$: these can be considered as statistical anomalies, and if the model perfectly captures the generating process, this should happen $5\%$ of the times on average; these are false positives. The second case corresponds to the anomalies that are artificially added, and can be considered as malfunctions, or true positives. 

In every experiment, we repeat the whole process (data generation, learning and anomaly detection) $N=10$ times to have an idea of the variability of the results.\\

\indent\textbf{Asymptotic setting}
In this setting, similarly to popular services on AWS, we will suppose we have access to a grid of a thousand quantiles of $F_t$ at each time step, and construct $\widetilde F_t$ from these quantiles. For this setting, we take a regular grid of $d=30$ bins, which is enough to detect the malfunctions with a high accuracy. We report the results in Tables \ref{tab:asymp_ds1_ds2}. The proportion of statistical anomalies is computed on the set of time points that are not malfunctions. This corresponds to the False Positives Rate (FPR). We can see that the proportion of such anomalies is stable around $5\%$, as it should. The introduction of $3\%$ of malfunctions does not seem to impact it too much. We can also notice that for dataset 1, where noise is added on the mean in the learning phase generating process, the algorithm is slightly worse at detecting distributional shift. And similarly, in dataset \ref{noise_var} where noise is added to the standard deviation, we see decreased performance in detecting distributional collapse. \\

\begin{table}[tb]
    \centering
    \begin{tabular}{lcc}
        \toprule
         & \textbf{False Positive Rate} & \textbf{Recall} \\
        \midrule
        \textbf{DS1 }  & $5.73 \pm 0.61$ & - \\
        \textbf{DS1 $\mu$}   & $5.60 \pm 0.99$ & $99.7 \pm 0.67$\\
        \textbf{DS2 $\sigma$}  & $5.43 \pm 0.11$ & $100$\\
        \midrule
        \textbf{DS2}  &$4.96 \pm 1.0$ & - \\
        \textbf{DS2 $\mu$}   & $5.15 \pm 1.5$ & $100$\\
        \textbf{DS2 $\sigma$}   & $4.98 \pm 0.72$& $99.8 \pm 0.5$\\
        \bottomrule
    \end{tabular}
    \caption{Anomaly detection for synthetic datasets in the asymptotic setting with our proposed method. Results are expressed in percent. When the name of the dataset is followed by $\mu$ (resp. $\sigma$), it corresponds to distributional shifts malfunctions (resp. distributional collapse). Otherwise, no malfunctions are introduced. We expect the FPR to be $5\%$ in all cases.}
    \label{tab:asymp_ds1_ds2}
\end{table}

\indent\textbf{Finite $n_t$ setting}
In this setting, we observe $n_t=60$ samples from evey distribution $F_t$, which corresponds to hourly aggregation of minute data. Here the task is more complex. We take a quantile grid of $d=10$ bins. With such a small dimension, the model can not always capture the changes in the trend, however, the relatively small amount of data prevents us taking a larger $d$, contrary to the asymptotic case.  In most practical settings, we are able to take $d$ much larger since we can make use of multiple time series simultaneously, even though they represent different metrics (CPU usage, Latency, Number of connected users, etc.). But for this synthetic setting, we restrict our method by learning and performing anomaly detection one time series at a time. 

The objective of this experiment is to see how well our approach performs compared to the standard approach of monitoring an aggregated statistic. We use two state-of-the-art open source anomaly detection algorithms, namely Luminol and TwitterAD, as competitors. These algorithms are run on the appropriate aggregated statistics. For the mean malfunctions we use the time series of empirical means (per hour), and for the variance malfunctions we use the empirical standard deviations. While we use the right aggregated statistic for the injected anomalies, we note that in a practical setting we don't know which statistics is most appropriate to monitor.

The results are reported in Table \ref{tab:small_ds_auc}. From this experiment we see that in general it is more complicated to detect standard deviation change malfunctions, with the worst case being dataset 2 with collapse malfunctions. The most plausible interpretation is that high order statistics are more difficult to estimate with a finite number of samples.

\begin{table}[tb]
    \centering
    \begin{tabular}{cccc}
        \toprule
         & \textbf{Distribution} & \textbf{TwitterAD} & \textbf{Luminol}  \\
        \midrule
        \textbf{DS1 $\mu$}  & 0.9928 & \textbf{0.9998} & 0.9400 \\
        \textbf{DS1 $\sigma$}  &  \textbf{0.9864} & 0.5010 & 0.9691\\
        \textbf{DS2 $\mu$}  &  0.9973 & \textbf{0.9999} & 0.9596 \\
        \textbf{DS2 $\sigma$}  &  \textbf{0.9797} & 0.4990  & 0.9456 \\
        \bottomrule
    \end{tabular}
    \caption{Comparative evaluation of anomaly detection methods on the synthetic high frequency data. Average AUC for $10$ simulated data. When the name of the dataset is followed by $\mu$ (resp. $\sigma$), it corresponds to distributional shifts malfunctions (resp. distributional collapse) }
    \label{tab:small_ds_auc}    
\end{table}

\subsection{Yahoo webscope Dataset}
Yahoo Webscope is an open dataset often used as a benchmark for anomaly detection since it is labeled. It is composed of 367 time series, varying in length from 700 to 1700 observations. Some of these time series come from real traffic to Yahoo services and some are synthetic. The dataset is divided into 4 sub-benchmarks, from $A1$ to $A4$.  The time frequency of all the time series is one hour. Since the frequency is relatively low, and since there are no collective anomalies in this dataset, we take $n_t = 1$. This setting corresponds to the regular anomaly detection setting, where we only have one observation per time step. We report the results of \cite{fuseAD} to compare the performance of our approach with the state of the art anomaly detection algorithms. We report the results per sub-benchmark, since they contain different patterns. We use  $40\%$ of each time series for training. We learn a single model for all the series of a same sub-benchmark, which means that we train the model on all the time series simultaneously. The hyper parameters are selected based on a few runs on the training set (which has been subdivided into training and validation sets).  The results are given in Table \ref{tab:yahoo_auc}. Here, since $n_t=1$, the total number of possible outcomes is equal to $d=100$. Therefore, we do not need Monte Carlo estimates.

\begin{table*}[t]
    \centering
    \begin{tabular}{ccccccccc}
        \toprule
         \textbf{Benchmark} & \textbf{iForest} & \textbf{OCSVM} & \textbf{LOF} & \textbf{PCA} & \textbf{TwitterAD} & \textbf{DeepAnT} & \textbf{FuseAD} & \textbf{Distribution}  \\
        \midrule
        \textbf{A1}  &$0.8888$ & $0.8159$ & $0.9037$ & $0.8363$ & $0.8239$ & $0.8976$ & \textbf{0.9471} & 0.9435\\
        \textbf{A2}   & $0.6620$ & $0.6172$ &	$0.9011$ &	$0.9234$ &	$0.5000$ & $0.9614$ & $0.9993$ & \textbf{0.9999}\\
        \textbf{A3}   & $0.6279$ &	 $0.5972$ &	$0.6405$ &	$0.6278$ &	$0.6176$ &	$0.9283$ & \textbf{0.9987} & \textbf{0.9988} \\
        \textbf{A4}   & $0.6327$ & $0.6036$ & $0.6403$	& $0.6100	$ & $0.6534$ & $0.8597$ & $0.9657$ & \textbf{0.9701}\\
        \bottomrule
    \end{tabular}
    \caption{Comparative evaluation of state-of-the-art anomaly detection methods on the Yahoo Webscope dataset. Average AUC per benchmark. Except the last column, the results of this table are taken from  \protect\cite{fuseAD}}.
    \label{tab:yahoo_auc}    
\end{table*}

\subsection{AWS data}
Finally, we consider three benchmark datasets of high frequency time series, collected from AWS. These datasets are often used internally at Amazon to compare models. The benchmark \textbf{B1} has a $1$ minute time frequency, it is composed of $55$ time series. The benchmarks \textbf{B2} and \textbf{B3} have a $5$ minute  time frequency. They are composed of $100$ time series each. All datasets are composed of different metrics, among them CPU usage, latency, number of users, and so on. Each time series of all three benchmarks have approximately $17\,000$ time points. We use $60\%$ of the time range for training and and the remainder for detection. We set $d=100$ and aggregate all time series to a $30$ minutes frequency, so $n_t = 30$ for \textbf{B1} and $n_t = 6$ for \textbf{B2} and \textbf{B3}. However the quality of the labeling is heterogeneous, \textbf{B1} being the most reliable one.  Some examples of inaccurate labeling from the third benchmark can be found in Figure \ref{plots:bad_labels} in the appendix. 
We find that the anomalies identified by our method are false positives under the labels but should probably be counted as true positives. We perform a two stage anomaly detection, the first stage gives scores for the single observations, the second for the collection of observations within half-hours. We simply add the two scores to get the final score. We again compare to Luminol and TwitterAD. The results are reported in Table \ref{tab:aws_auc} which show the dominance of our method on this data set.

\begin{table}[tb]
    \centering
    \begin{tabular}{cccc}
        \toprule
         & \textbf{Distribution} & \textbf{TwitterAD} & \textbf{Luminol}  \\
        \midrule
        \textbf{B1}  & \textbf{0.8183} & 0.7134 & 0.6467\\
        \textbf{B2}  &  \textbf{0.7534} & 0.5895 & 0.5804\\
        \textbf{B3}  &  \textbf{0.6860} & 0.5889 & 0.5860\\
        \bottomrule
    \end{tabular}
    \caption{Comparative evaluation of anomaly detection methods on AWS data. Average ROC-AUC per benchmark.}
    \label{tab:aws_auc}    
\end{table}

\section{Related work}\label{sec:related_work}

Anomaly detection is a rich field with many applications and solutions available (see for example~\cite{anom1, anom2, anom3, anom4, anom5, schelter18,cleanits,twitter,microsoft,timeseriesCleaning}). Since we consider a specific anomaly detection method on time series here, we focus our discussion of related work on \emph{methods} in the following (as 
opposed to systems or solutions) and even more specifically, unsupervised anomaly detection models. Such methods are appropriate in our scenario because 
labels in industrial settings as we consider them with millions of time series, are sparse. However, we acknowledge that supervised and semi-supervised methods 
would be an interesting avenue for further work~\cite{gao2020robusttad}. 

Most related to our approach are anomaly detection methods using (probabilistic) time series or forecasting models. Overview of forecasting models 
can be found in recent tutorials~\cite{Faloutsos2018,Faloutsos2019}. These approaches 
have the advantage that we can obtain an interpretable and normalized score (e.g., how likely is a point under the model) which allows to tune thresholds over large panels 
of time series simulatenously. Other approaches, e.g.,~\cite{rcf-guha16,matrixprofileI}
do not allow for this and hence we do not consider them further. 

The anomaly detection algorithm consists in estimating the parameters of a time series or forecasting model in a first step. Then, the score of a point in the time series will be a quantification of the distance between the prediction and the observation (for example $\ell_2$ distance or the likelihood of the observation if the model is probabilistic). The choices of the time series model differ in the literature. The most common models are the classical ARIMA (see for example \cite{arima1,arima2} for applications) and increasingly deep learning based approaches as we apply it here~\cite{deep1,deep2,fuseAD}. Even though in the more general area of sequence learning, attention-based models~\cite{vaswani2017attention} have become the state of the art, our choice of an RNN similar to~\cite{salinas2019deepar,gasthaus2019probabilistic} is motivated by the streaming setting that we consider here. 
The compact model state of an RNN is well suited to a streaming setting, whereas attention-based models have 
a prohibitely large state for a streaming setting. This is why we do not further consider them or convolutional alternatives~\cite{van2016wavenet} here. 

Existing time series models that consider distributional data rely on Functional Data Analysis to provide the necessary mathematical tools for analyzing such problems (see \cite{ramsay2004functional}, \cite{horvath2012inference} or \cite{ferraty2006nonparametric} for an overview of existing non parametric methods). These Functional Time Series (FTS) models are the most related to our problem. In that setting, the learner observes a time sequence of functional data, and tries to forecast the next functions. This framework and the resulting methods have many applications ranging from the study of demographic curves (for example \cite{hyndman2007robust,hyndman2009forecasting}) to electricity price forecasting (see \cite{gonzalez2017forecasting}). While these models allow for more general functions than distributions as data points, the 
restriction to distributions has led to further models~\cite{park2012functional,chang2016nonstationarity,chang2019evaluating,koko2019forecasting} and Bayesian variants 
have also been proposed~\cite{caron2007bayesian,rodriguez2008bayesian,mena2016dynamic}. 

Instead of time series models, it is also possible to disregard time dependencies and case the time series prediction as a regression problem. This would allow for employing distribution regression models such as~\cite{muandet2012learning,law2018bayesian,szabo2016learning,poczos2013distribution,oliva2013distribution,kou2019compact}. Given the strong auto-correlation in metrics data as we observe in our application, time dependence should not be disregarded and we therefore do not consider this approach further. To the best of our knowledge, we are not aware of other anomaly detection methods relying on distributional time series models beside ours.

\section{Conclusion}\label{sec:conclusion}

We presented the first anomaly detection method based on deep distributional time series models. The development of this model was motivated by real-world anomaly detection data and use-cases that we commonly find in monitoring cloud services. In the experiments, we show that on synthetic, public and AWS-internal data, our method compares favorably to other 
anomaly detection offerings. Our method was designed for streaming scenarios as they occur in monitoring compute metrics and it is fully elastic.

While labels for anomalies are sparse, imbalanced and noisy, they nevertheless exists. Future work should consider how to 
improve the algorithms described here by incorporating labels during learning and acquiring them during production runs 
to lead to a continuously improving anomaly detection system.

\bibliographystyle{unsrt}

\newpage
\section*{Appendix}

\subsection{RNN architecture:}
\begin{figure}[h]
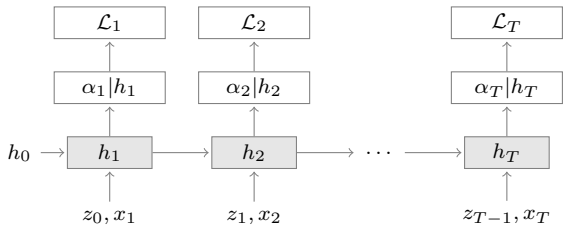

   \begin{minipage}{0.48\textwidth}
     \centering
     \networkTrainingPictureSmall
     \caption{Training: At each time step $t$, the inputs to the network are the covariates $x_{t}$, the target value at the previous time step $z_{t-1}$, as well as the previous network output $h_{t-1}$. The network output $h_{t} = r_\phi(h_{t-1}, z_{t-1}, x_{t}))$ is then fed to the projection layer, which outputs the parameters $\alpha_{t} = \theta_\phi(\mathbf{h}_{t})$. Finally, the negative log-likelihood of the observation given $\alpha_t$ is used as loss to train the model.}\label{fig:training}
   \end{minipage}
\end{figure}

\begin{figure}[h]
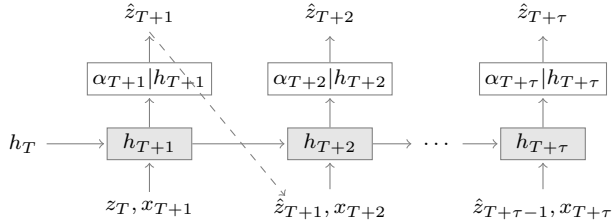

   \begin{minipage}{0.48\textwidth}
     \centering
     \networkInferencePictureSmall
     \caption{Prediction: In the conditioning range ($t\leq T$) the (known) history of the time series $z_{t}$ is fed in to the network, along with the corresponding covariates. In the prediction range ($t>T$) a sample $\hat{z}_{t}$ given $\alpha_t$, is drawn and provided as input for the time step, until the end of the prediction range $t=T+\tau$, generating one sample path. Repeating this prediction process yields samples from the joint prediction distribution.}\label{fig:prediction}
   \end{minipage}
\end{figure}

As explained in the main text, we will use an autoregressive LSTM-based recurrent neural network whose architecture follows the one described in \cite{salinas2019deepar}. The Figures \ref{fig:training} and \ref{fig:prediction} hereafter give a high level description of the model's architecture for the training and prediction steps. Further details can be found in the aforementioned work. In an anomaly detection setting, we only consider one step ahead prediction, and therefore take $\tau = 1$ in the prediction network. 

\subsection{Further experimental results}

\subsubsection*{Synthetic data: Asymptotic setting}
\begin{figure}
    \centering
        \includegraphics[scale=.35]{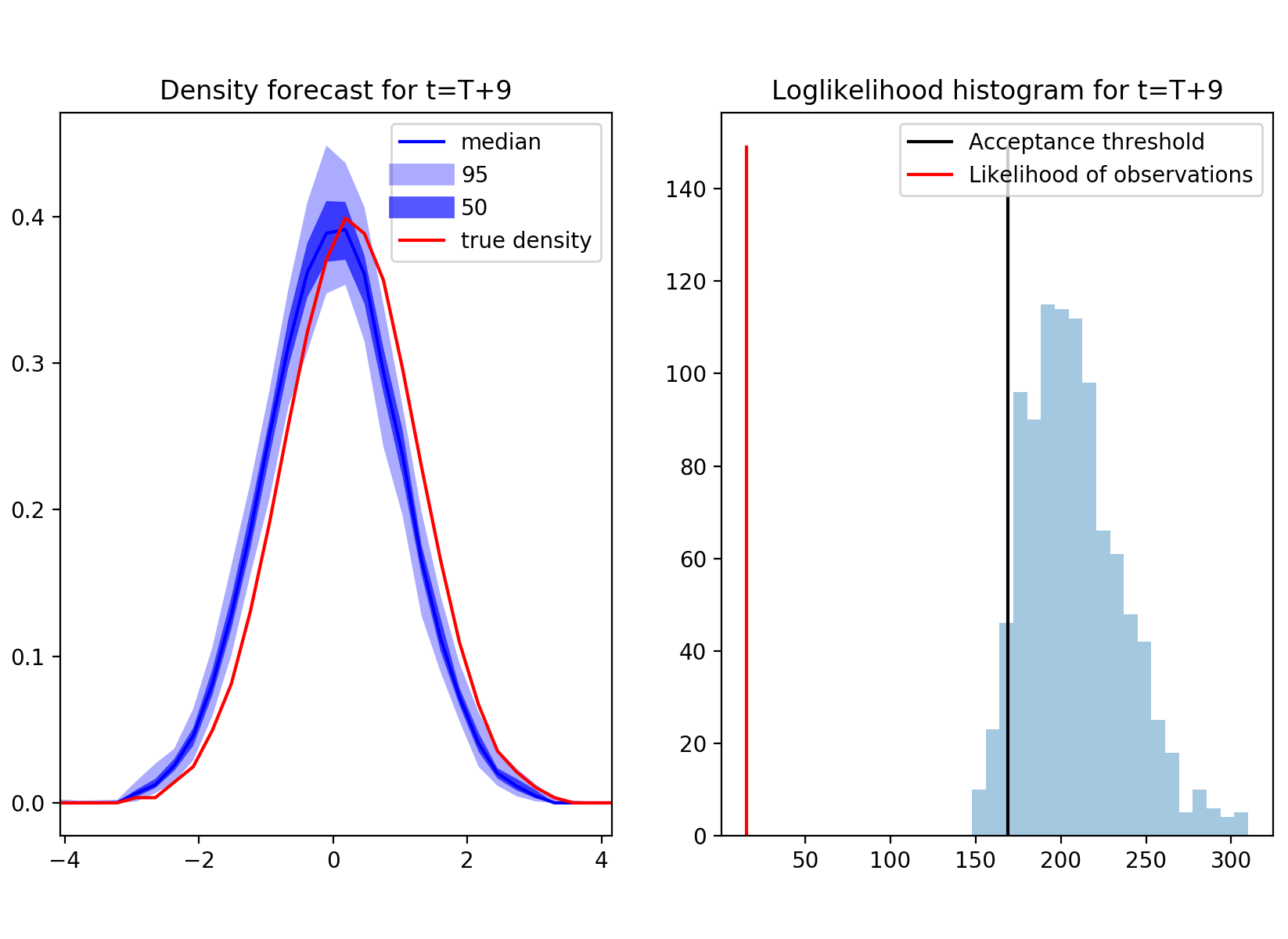}
        \includegraphics[scale=.35]{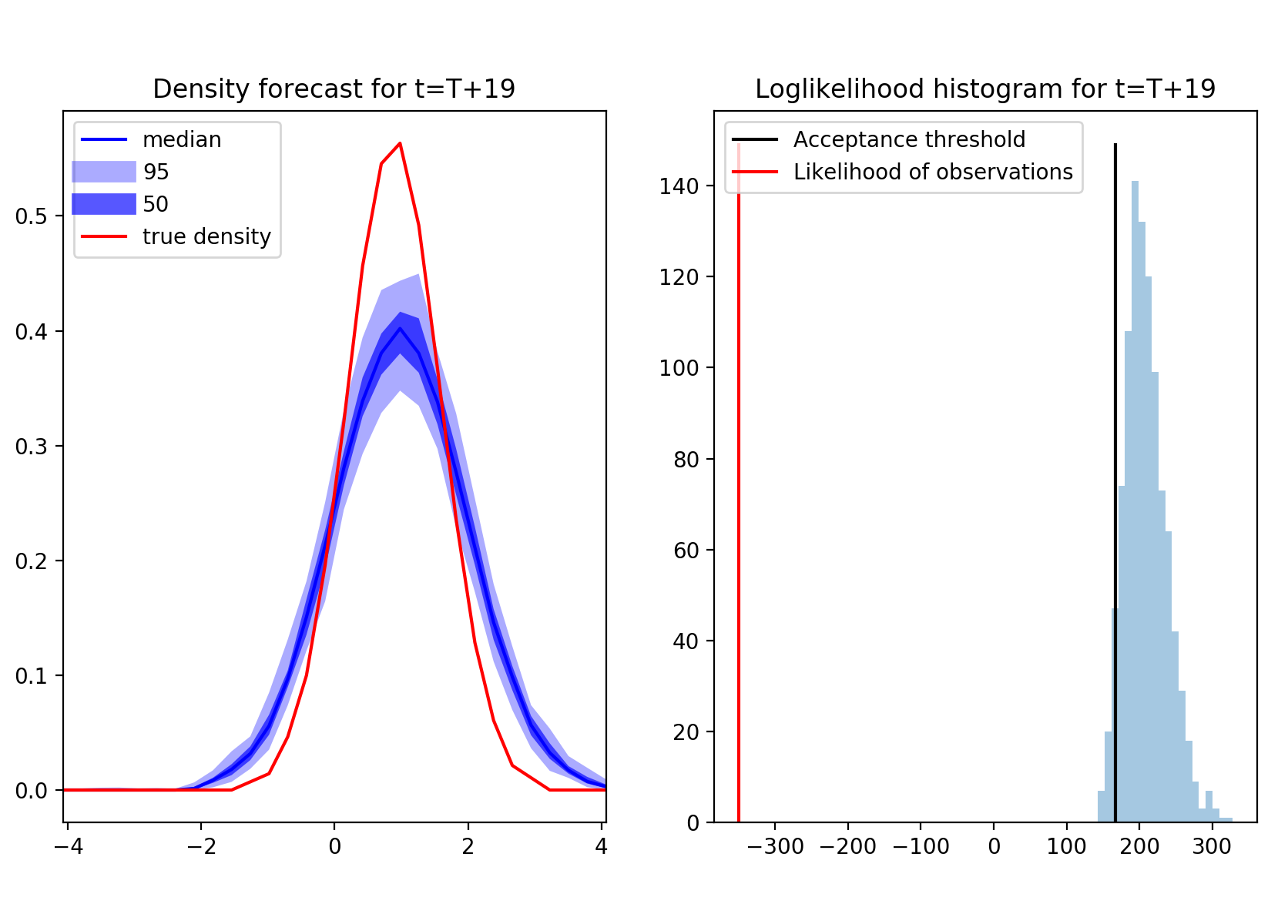}
        \includegraphics[scale=.35]{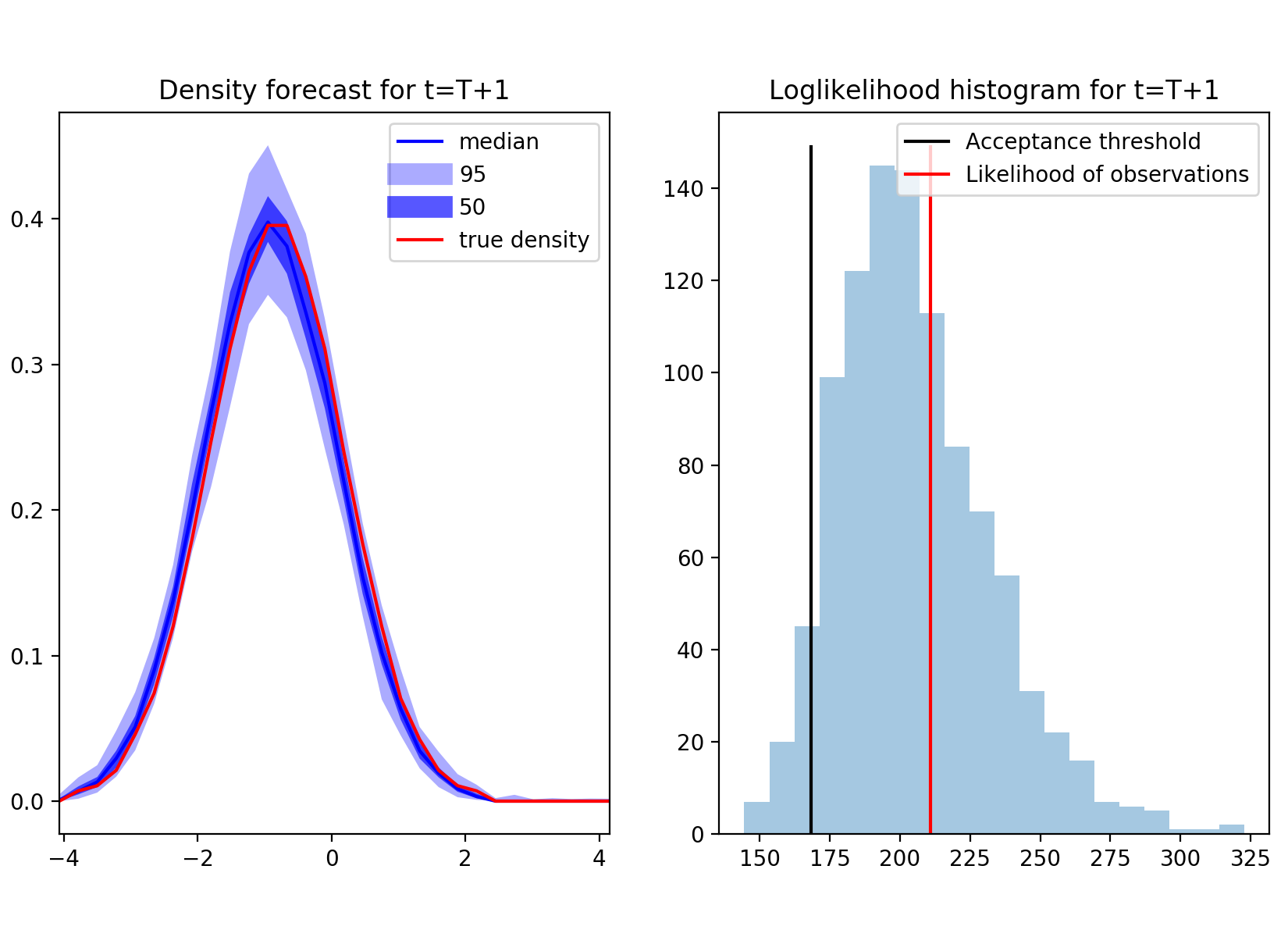}
    \caption{Examples of anomalies for the dataset \ref{noise_mean} where the malfunctions consist in reducing the variance of the distributions. (Top) Statistical anomaly. (Middle) Malfunction anomaly. (Bottom) Non anomalous distributional observation. For the three different dates, we show: on the left figures, the observed distribution (in red) and the coordinate-wise marginal credible intervals; on the right figures,  the histograms of log likelihoods where the red line corresponds to the log likelihood of the observed distribution and the black line the acceptance threshold. }\label{plots:simulated}
\end{figure}

Figure \ref{plots:simulated} show an example of a statistical anomaly as well as a malfunction for the dataset 1 with collapse malfunctions in the asymptotic setting: we show the observed distribution (in red), the predicted coordinate-wise credible intervals (blue). We also show the predicted distribution of log-likelihoods, using the histogram of the $M=1000$ samples from the predictive.

\subsubsection*{Synthethic data: Finite $n_t$}
In Table \ref{tab:finite_ds1_ds2}, we report the average False Positives Rate and Recall in the finite $n_t$ setting.

\begin{table}[h]
    \centering
    \begin{tabular}{lcc}
        \toprule
         & \textbf{False Positives Rate} & \textbf{Recall} \\
        \midrule
        \textbf{DS1 }   & $5.45 \pm 0.72$ & - \\
        \textbf{DS1 $\mu$}   & $5.30 \pm 0.72$ & $97.7 \pm 2.2$\\
        \textbf{DS2 $\sigma$}   & $4.65 \pm 0.70$ & $95.1 \pm 3.8$\\
        \midrule
        \textbf{DS2}  & $5.33 \pm 0.92$ & -  \\
        \textbf{DS2 $\mu$}   & $4.70 \pm 1.1$ & $98.5 \pm 1.1$\\
        \textbf{DS2 $\sigma$}   & $5.27 \pm 1.5$& $91.7 \pm 3.3$\\
        \bottomrule
    \end{tabular}
    \caption{Anomaly detection for synthetic datasets in the finite $n_t=60$ setting with our proposed method. Results are expressed in percent. When the name of the dataset is followed by $\mu$ (resp. $\sigma$), it corresponds to distributional shifts malfunctions (resp. distributional collapse). Otherwise, no malfunctions are introduced. We expect the FPR to be $5\%$ in all cases.}
    \label{tab:finite_ds1_ds2}
\end{table}

\subsubsection*{AWS data}
\begin{figure}[h]
    \centering
        \includegraphics[scale=.3]{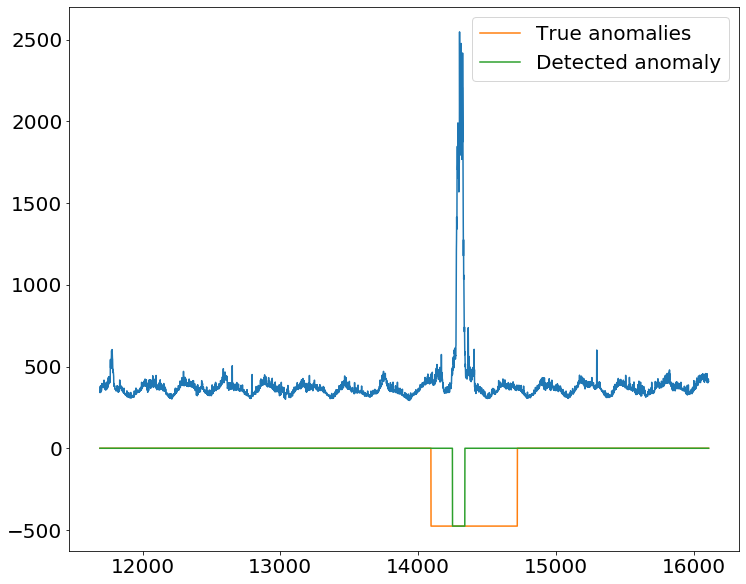}
        \hspace{5em} \includegraphics[scale=.3]{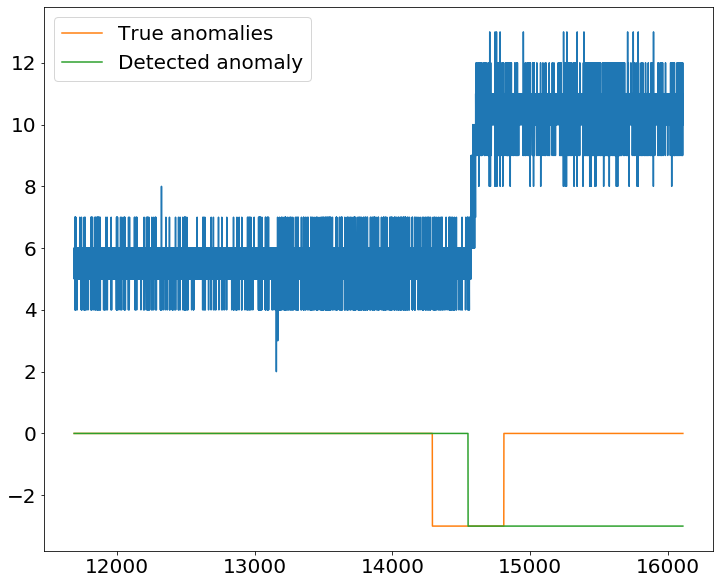}
    \caption{Examples of labeled time series from AWS benchmark \textbf{B2}. The yellow lines represent the points labeled as anomalies, and the green line the points detected as anomalies by our method}\label{plots:bad_labels}
\end{figure}

In Figure \ref{plots:bad_labels} we give two examples of time series from the AWS \textbf{B2} benchmark dataset. As we can see, the labeling can be inaccurate which explains why the scores of all methods are not very high.


\end{document}